# Designing an AI-Driven Talent Intelligence Solution: Exploring Big Data to extend the TOE Framework

**Ali Faqihi and Shah J Miah**

Newcastle Business School, University of Newcastle, NSW, Australia

**Abstract**

AI has the potential to improve approaches to talent management enabling dynamic provisions through implementing advanced automation. This study aims to identify the new requirements for developing AI-oriented artifacts to address talent management issues. Focusing on enhancing interactions between professional assessment and planning attributes, the design artifact is an intelligent employment automation solution for career guidance that is largely dependent on a talent intelligent module and an individual's growth needs. A design science method is adopted for conducting the experimental study with structured machine learning techniques which is the primary element of a comprehensive AI solution framework informed through a proposed moderation of the technology-organization-environment (TOE) theory. Moreover, the study will recommend practical directions for an IS researcher for addressing unique talent growth requirements in the organization, in addition to automated features in talent management solutions.

## 1. Introduction

The use of artificial intelligence (AI) in talent management has been already established as a significant research trend. Many modern technologies address the issues involved in developing such systematic automated information support solutions, but AI has been rarely applied for enhancing practices in employment management (Vrontis et al., 2022). Capabilities of AI are viewed in existing cases of studies for the construction of interventions in employment prospective, but various disruptive innovations to enhance the current frameworks of talent systems are not holistically studied in the past recent years. The advancement of general-purpose AI technology is of paramount task to revolutionizing workforce management (Agrawal et al., 2018). While creating new AI oriented applications for employment management, a number of obstacles such as dehumanization, biased algorithms and fairness in requirement have identified, so it is imperative to conduct precise design research (Tambe et al., 2019).

A recent industry survey identified at least 300 HR technology start-ups developing AI tools for people management, with roughly 60 of these companies achieving traction in terms of clients and venture investment (Bailie & Butler, 2018). Furthermore, an AI-powered talent intelligence platform that aids in attracting, developing, and retaining outstanding employees, has just raised $220 million and is now valued at over $2 billion (Charlwood & Guenole, 2022).Many organizations have started with their massive investment in AI for workforce management. HR technology startup that automates work processes in order to improve many practices such as, the job search experience, has *"quietly become a billion-dollar unicorn"* (Kelly, 2021). Accenture has invested strategically in the London-based startup Beamery, which provides options for their recruiting operating system and is valued at $800 million (Lunden, 2021).Large multinational technology organizations have already begun integrating AI into their talent acquisition systems and procedures (van den Broek et al., 2021) IBM calculated that the cost savings from implementing AI in HR were close to $100 million per year (Guenole & Feinzig, 2018).This implies that talent management would be an important part to AI initiatives in organization indicating research significance in the field.

Talent management is an area of HR operations in which various data about individuals' skills and experience, as well as the skills required for certain jobs, are managed; however, such huge amounts of workforce employment data are divisive in nature and huge in volume and rapidly evolving in terms of their complexity (Collings & Mellahi, 2009).A new technological approach using AI can address the issue of managing such data by transforming them into useful insights for both managerial and operational decision-making. In this paper, we propose a new talent management AI approach that utilizes machine learning to combine different data sources for delivering more insightful predictions. The use of AI reveals issues, such as discrimination, biased algorithms, and dehumanization (Charlwood & Guenole, 2022; Fritts & Cabrera, 2021). Therefore, when designing and evaluating the AI solution, it is important to consider both the potential benefits and risks of designing such systems. In particular, decision-makers should ensure that AI systems are designed in a way that minimizes the risk of perpetuating existing biases and discriminatory (Shrestha et al., 2019). Furthermore, it is important to monitor AI based solutions on an ongoing basis to ensure that they are functioning as intended morally with an attachment of human ethics and not causing any negative unintended consequences.

The remaining portion of the article is organized as below. The essential background literature is discussed in Section 2, which provides an overview of the solution artefact design. The methodology of the research study is broken down in depth, and the next part delves into the specifics of the design artefact, situating our contribution within the framework of earlier work that is pertinent to the topic at hand. Following the part on the discussion, which focuses on the general contributions of the study, comes the section on the conclusion.

## 2. Background of the study

AI helps organizations to manage large volumes of data more effectively, as well as to identify patterns and extract insights that would be difficult to discern using traditional methods (Nemati et al., 2002). In the context of talent management, AI can be used to improve the accuracy of job candidates' assessments, as well as to identify potential employees who may be a good match for vacant positions (Jia et al., 2018). Therefore, AI oriented management solutions have the potential to improve organizational efficiency by reducing the time and resources required to screen job applicants and identify suitable candidates.

Despite the potential benefits of AI in HR, there are also risks that need to be considered. For example, AI-based systems may offer biases and discriminatory practices (Chamorro-Premuzic et al., 2019).When designing and evaluating AI-powered talent management solutions, it is therefore important to consider both the potential benefits and risks of using such systems. In particular, decision-makers should ensure that AI systems are designed in a way that minimizes the risk of perpetuating existing biases and discriminatory practices. When compared to various other HR procedures, talent management is one in which the impact of AI can be seen as a driving force for substantial improvements in terms of the practices. However, if AI is to be applied in this process, we will need to effectively outline a conceptual solution model to meet its key requirements. Following table 1 shows existing studies in the problem domain.

Table 1: Existing studies in AI for talent management

| *Studies related to AI applications in talent management* | *Used ML approaches* | *Key findings of the papers* |
|---|---|---|
| Fritts & Cabrera (2021) | ML concepts identification. | Examines the issue of recruitment algorithms with an eye toward the under-explored concerns of HR managers. |
| Xiao & Yi (2021). | Tensorflow platform for supervised ML | Implements the design using AI to career planning or related areas. |
| Joshi, Goel & Kumar (2020). | Support Vector Machine (SVM). | Builds AI solutions for career services. |
| Zhao et al. (2021). | Algorithm model design | AI for addressing fairness concerns for designing recruitment systems |
| Beloff & White (2021). | a minimum viable product (MVP), Natural Language Processing (NLP), and explanatory knowledge derived system. | AI architecture that holds AI models and a data repository for recruiting model |
| Shafagatova & Van Looy, (2021) | Supervised ML | AI for "process-oriented appraisals and rewards" |
| Meng (2017) | Supervised ML | Utilizes the design routine of modular design, real-time evaluation, and standard analysis for assessing emotional stability |

**2.1 Ethical Implications of AI in Talent Management**

Many studies have addressed the need for AI in improving workforce processes. The promise and reality of AI in workforce management are vastly different to acquire, manage and retain talent- where there are complexity, constraints of data set and fairness (Tambe et al., 2019). In order to rapidly improve talent management enabling advantages of the potential of various data sets through the application of AI tools, it is important to design information systems solution that will support shifting organizational focus from developing more ethical HR systems for better efficiencies (Chamorro-Premuzic et al., 2019). This implies that the need of ethical AI solutions in Talent Management is essential to govern the hiring and selection process. Uncertainty in the labor market, fast-changing business strategies, and new technology in workforce operations all call for a dynamic talent management system. Numerous scientists have expressed a variety of ethical concerns regarding AI such as predictive analytics algorithms to make judgments that have a significant impact on the life chances of humans (Mittelstadt et al., 2016). It implies the demands of developing more practical solutions that would be meeting the demands of three overlapping workforce principles: causal reasoning, randomization and experiments, and employee contributions (Tambe et al., 2019).

**2.2 The Application of AI in Talent Management**

The fourth industrial revolution, driven by demography, technology, and globalization, has fundamentally altered employment, with far-reaching ramifications for workers. In recent years, AI has begun to be applied to the field of talent management. Talent management relates in improving process of identifying, developing, and retaining employees and their potential to contribute to an organization's success (Collings & Mellahi, 2009).Going beyond to automate repetitive tasks such as job postings and resume screening by AI applications, it is important to identify patterns in data that can help organizations to make better decisions about talent development and retention.

The use of AI in career services is still in its early stages, but there is potential for it to be used in a variety of ways. For example, AI can be used to create predictive models that can identify which students are at risk of not getting a job after graduation (Mehraj & Baba, 2019). A number of AI-based tools have been developed for career guidance purposes. One such tool is Job Finder, which uses machine learning to match job seekers with appropriate jobs based on their skills and preferences (Liu et al., 2017).There is no doubt that AI has the potential to assist career counselors with some of their daily tasks. For instance, it can help in resume writing and interview preparation. Additionally, AI can be used to create a bot that helps students search for jobs and internships. The bot can also provide personalized recommendations based on the student's interests and skills. Moreover, AI can be used to develop a chatbot that can answer frequently asked questions about careers, job applications, and resumes. Based on the above discussion, we identified a requirement for developing a new AI driven talent management solution. AI Talent Intelligent will present with innovative options to integrate technology in career advice which result in enhanced cost-effectiveness, decreased total expenses, and higher cost-effectiveness, in addition to improvements in accessibility and increased access to information, assessment, and networks (Sampson et al., 2020).

### 3. Design Science Methodology

This study aims to design a new AI solution therefore it is essential to investigate its suitability in the design context. We plan to develop a methodological framework for the deployment of talent management AI software solutions for career services. Key components of talent management and related processes integrating talent management systems will be extracted from the latest relevant literature and validated with secondary data collected from social media or any other sources that make big training and development datasets freely available. The emerging themes of the proposed proof-of-concept prototype (e.g. Miah, 2008; Miah, 2010; Miah, Vu, & Gammack, 2018) will also be designed and compared with existing software solutions available in the market for talent management. A comprehensive talent management artefact will be prototyped, which will meet the concurrent talent requirements and support the future operations of Career Centre to sustain their operations and help talents succeed in their future careers. Following table 2 shows the adopted design science research (DSR) principles adopted from Hevner et al. (2004).

Table 2: Hevner's seven guidance in a summarized form

| DSR Adopted Guideline | Its relevant to the proposed solution |
|---|---|
| Guideline 1: Design as an Artifact: Design- | Design-science research must result in a valid construct, model, technique, or instance. |
| Guideline 2: Problem Relevance: | The objective of design-science research is to create technological response to significant business issues. Identified is a genuine issue domain that supports the specified software solution prototype. |

| Guideline 3: Design Evaluation: | The utility, quality, and feasibility of a AI design need to be stringently shown by well-executed assessment procedures in order to satisfy the requirements. For prototype testing with industry various stakeholders, a descriptive assessment approach will be used with utilizing secondary data. |
|---|---|
| Guideline 4: Research Contributions: | The models utilised for the AI artifact's features were designed by domain specialists with information gleaned from actual practice, through prototyping. |
| Guideline 5: Research Rigor: | DSR is dependent on the use of rigors procedures in the creation and assessment of the AI design artefact informing through IS theories. |
| Guideline 6: Design as a Search Process: | The search for a functional artefact necessitates the utilisation of accessible ways to achieve desired purposes in compliance with the issue domain, |
| Guideline 7: Communication of Research: | DSR will help successfully communicated about the research outcome to both technical and managerial groups. |

Hevner et al. (2004) described that DSR is a scientific approach to problem-solving that emphasizes the development of new knowledge to design and evaluate solution artifact for addressing real-world problems. The main goals of DSR are to: 1) developing new artifacts or software system model, method or construct for real-world problems; 2) evaluating effectiveness of these solution artifact and 3) generating new knowledge about the design of new systems solution. Following an ensemble artefact design (Miah & Gammack, 2014), the first step in this process is to develop a conceptual model that describes how AI can be used to improve talent management. This model will be based on a review of the literature on AI and talent management. Next, the feasibility of using AI in talent management will be assessed, through datasets. This will involve conducting reports in the field to identify the challenges and opportunities associated with AI-based solutions. Finally, a prototype AI-based solution will be developed and tested in two problem contexts (e.g. using design principles defined by Miah, Gammack & McKay, 2019). The results of this study will contribute to our understanding of how AI can be used to improve talent management.

In this study, secondary data will be employed to accomplish the study's purpose. The study will use a mapping approach to locate and assess AI-enabled companies, corporations, and/or initiatives that have begun to look for solutions to solve the cognitive and employability gaps. The steps of the search process used to produce an engaging map are outlined below:

1. The investigation of professional growth and employability frameworks.

2. The exploration of research on AI applications in TM and employability and associated with higher education.

3. Desk research on the uses of AI in recruiting and those associated with career centers.

4. The mapping of AI-using organizations within the framework for TM and employment.

5. The utilization and analysis of a few programs to develop a greater understanding of how they function.

6. Finally, the analysis of the collected data and the development of an innovative talent employability framework that leverages AI's ability to learn, identify complex patterns, and provide more accurate insights to assist students in making more informed career decisions, universities in attracting and serving students, and employers in selecting talent and accelerating the hiring process.

This study will employ an integrated framework based on the technology, organization, and environment (TOE) framework as well as on the diffusion of innovation (DOI) theory. The research investigates the significant aspects that are relevant for the adoption of AI TM artefacts in the career services of the higher education sector. The technology viewpoint of the TOE model takes into account the external as well as the internal technical resources that are necessary for the implementation of technology in an organization (Depietro et al., 1990).The technology factors discussed are reliability, security, capability, quality, relative cost advantages, and the compatibility of IT solution and technology benefits (Al-Qirim, 2006).

4. **Proposed AI oriented TM approach**

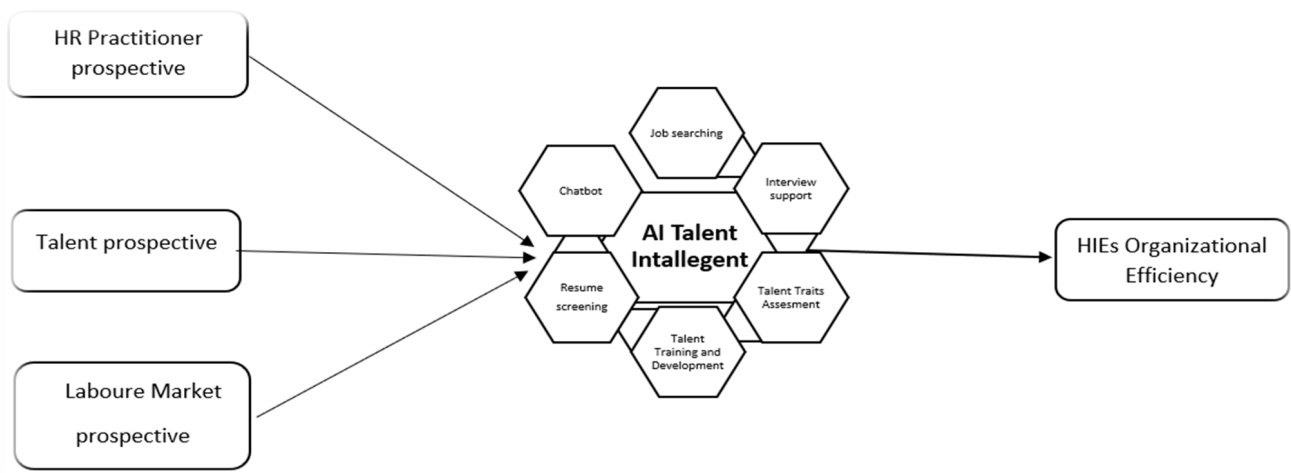

*Figure 1: Proposed Conceptual Framework*

Figure 1 shows a concept that we have drawn from the existing literature. An evolving prototype is an innovative design approach for transforming system requirements into a reasonable solution model design so that potential users may see the advantages or conceptual outcome. The understanding required to identify the problems with a present system and create an artefact solution is covered in the prototype design. Three phases are considered for the creation of the minimum viable product (MVP). These phases will include the following desired services: the talent pool, graduate personality traits assessment, career guidance chatbots, skills and knowledge building, CV screening software, mock interview screening, and jobs or opportunities matchmaking. The bots are software tools that have been created in natural language and have the ability to carry on conversations by themselves (Khan & Das, 2018). It is mostly divided into three different portions. The very first thing that must be done is to obtain information from the end-user. This can be done either vocally or by typing in natural language. The second step is to get the bot to provide output in the form of spoken words, and the third step is to process the input through the software in order to get output that is accurate and simple to comprehend. Most of the time, these counselling bots are utilized for producing predictions and questions regarding a career, such as which industry would be the best choice and which courses are the most recent. This system's principal function is to understand human speech with the assistance of natural language processing so that the results can be given to the user.

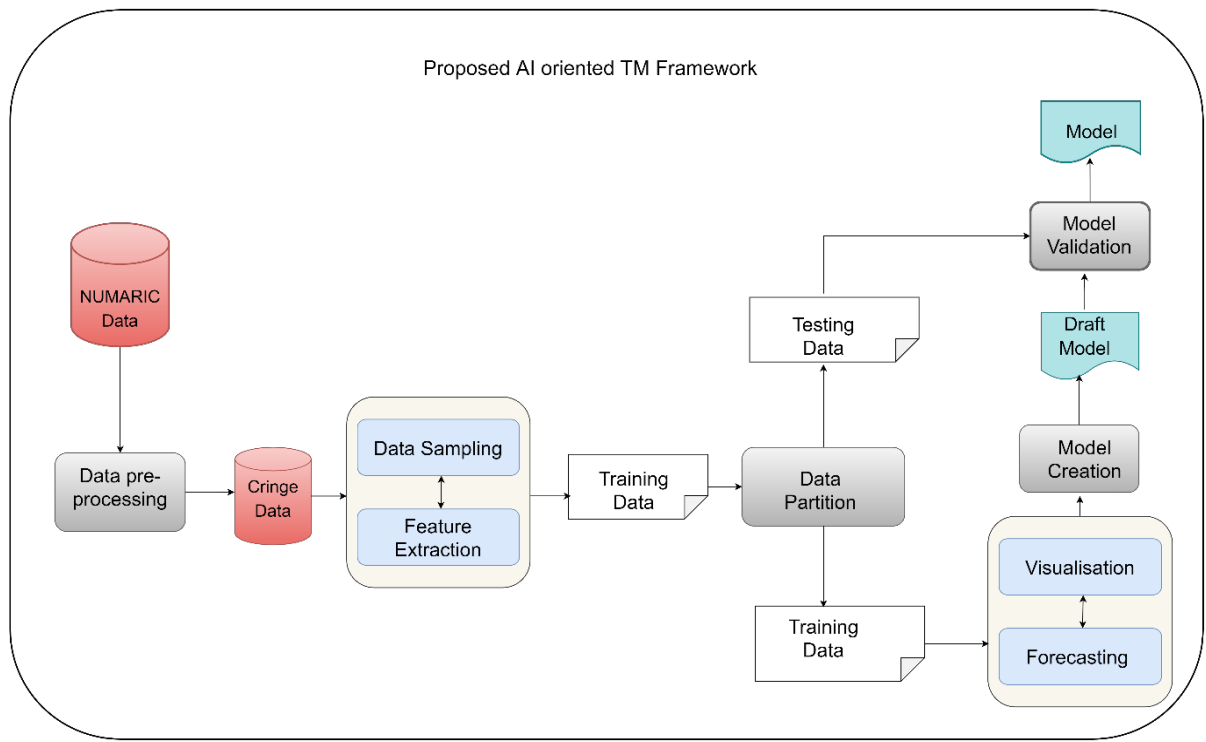

*Figure 2: A proposed AI Framework*

The proposed framework in figure 2 illustrates how the input (Numeric data) is going to convert to an AI-Driven Talent Intelligence Solution When our big data has been pre-processed (e.g., data cleansing, data integration, stop word removal), it is ready for sampling process and then be ready for separating into training and testing datasets. Figure 2 presents the proposed framework, which depicts how the output (an AI-driven talent intelligence solution) is going to be converted from the input (numerical data). When our big data has been pre-processed (for example, via data purification, data integration, stop word removal), it is ready for the sampling process, and after that, it will be ready to be separated into training and testing datasets. The generation of a draught model may be accomplished by the application of data analytics techniques such as visualization and forecasting. Following an analysis of its validity, the model will be transformed into a rigorous representation that may be put to work in forecasting, planning, and decision-making.

5. **Discussion and conclusion**

Artificial intelligence (AI) is hailed as a revolutionary, all-purpose technology that will transform the workplace. In the field of HR, AI is becoming increasingly prevalent. The proposed hybrid approach to TM will utilize multiple data sources to deliver data in a larger loop. We aimed to examine the design and development of a new AI-driven TM artefact. Our project will facilitate the daily operations of universities. AI can assist professional talent managers by identifying, forecasting, and investigating talent acquisition and performance-related trends. There is a growing demand for automated career counselling processes. Various sectors have adopted talent intelligence, including higher education systems, labour markets, the private sector, and the government. Higher education requires innovative ways to prepare students for the workforce. Smart technologies have the potential to assist in mentoring both current professionals and future talent.

Our aim is to create the AI platform that can help employment centre staff identify the best candidates for open positions, as well as to provide recommendations on how to improve organizational efficiency. Our case context is a university career center. In this project, we develop an initial machine learning algorithm that is trained by the data gathered on employment centers' current staffing levels, job openings, and hiring practices. Subsequently, the AI algorithm is to test a sample employment center data to determine, if it can accurately predict which candidates are best suited for open positions. Finally, The AI algorithm as the purposeful artifact is to be evaluated to determine its effectiveness in improving employment centre organizational efficiency.